\documentclass{article} 
\usepackage{iclr2027_conference,times}

\usepackage{amsmath,amsfonts,bm}

\def\eqref#1{equation~\ref{#1}}

\def\1{\bm{1}}

\DeclareMathAlphabet{\mathsfit}{\encodingdefault}{\sfdefault}{m}{sl}
\SetMathAlphabet{\mathsfit}{bold}{\encodingdefault}{\sfdefault}{bx}{n}

\usepackage{url}

\usepackage{booktabs}
\usepackage{longtable}
\usepackage{array}
\usepackage{multirow}

\usepackage{arydshln}
\usepackage{enumitem}

\usepackage{wrapfig}
\usepackage{amssymb}
\usepackage{graphicx}

\usepackage{capt-of}
\usepackage{xcolor}

\definecolor{cvprblue}{rgb}{0.21,0.49,0.74}
\usepackage[colorlinks=true, citecolor=cvprblue,
linkcolor=blue,
anchorcolor=red,
urlcolor=SkyBlue]{hyperref}

\title{Same Target, Different Origins: Aligning Latent Rollout for Consistent PDE Emulation}

\iclrfinalcopy

\author{
\textbf{Xinying Li} $^{1,2, \dagger}$ \quad
\textbf{Jing Xu} $^{2}$ \quad
\textbf{Wenjun Wang} $^{1}$ \quad
\textbf{Limei Han} $^{1,2}$ \quad
\textbf{Fenglei Cao} $^{2}$ \\[0.3em]
\textbf{Ruoxi Jiang} $^{1,2,\dagger,*}$ \quad
\textbf{Yuan Qi} $^{1,2,*}$ \\[0.5em]
$^{1}$ Fudan University,
\quad $^{2}$ Shanghai Academy of AI for Science \\[0.3em]
 $^{\dagger}$ Project leader,
 $^{*}$ Corresponding author \\
 \texttt{\{roxie\_jiang,qiyuan\}@fudan.edu.cn}
}

\begin{document}

\maketitle
\fancyhead[L]{Preprint}

\begin{abstract}

Neural PDE solvers commonly generate multi-step predictions through autoregressive rollout, recursively applying a learned evolution operator and feeding each prediction back as input for the next step.
Approximation errors can therefore propagate through the rollout, causing predictions at a shared target time to depend on their temporal origins and rollout depths.
For a well-posed deterministic PDE flow, however, states at different times along the same physical trajectory should, under exact evolution, reach the same target state.
To this end, we propose \textbf{A}lign \textbf{L}atent \textbf{R}ollout ({ALR}), a regularization approach that promotes consistency across rollout origins through the neural PDE solver’s internal representations.
A second-order Taylor analysis of latent discrepancy propagation motivates two objectives: zeroth-order latent state consistency, which aligns representations at a shared target time, and first-order latent dynamics consistency, which aligns local evolution responses to shared perturbations.
ALR can be integrated into different neural PDE solvers by regularizing suitable internal representations, without modifying their backbone architectures.
Experiments across multiple PDE benchmarks and neural-operator
backbones show that ALR improves prediction accuracy while reducing
shared-target discrepancies across rollout origins.


\end{abstract}

\section{Introduction}

Partial differential equations (PDEs) provide fundamental mathematical models for spatiotemporal systems across science and engineering. However, repeated numerical simulation can be computationally expensive, motivating data-driven surrogates that approximate solution operators. 
Among the existing approaches, neural operators learn mappings between function spaces and have emerged as a general framework for parametric PDE modeling~\citep{li2020fno, lu2021learning,kovachki2023neural}. 
For time-dependent PDEs, neural solvers commonly approximate a discrete-time evolution and recursively apply it through autoregressive rollout~\citep{li2020fno,brandstetter2022message,lippe2023pde}. 
As each prediction is fed back as input, local approximation errors may propagate and accumulate over the rollout, degrading the prediction accuracy.

Existing approaches mitigate autoregressive error accumulation from several perspectives \citep{ruhling2023dyffusion, list2025differentiability, jiang2026hierarchical}.
Autoregressive or multistep training can optimize models over recursively generated states, improving their robustness during rollout~\citep{brandstetter2022message}. 
Solver-in-the-loop methods train learned components through repeated interaction with numerical simulation~\citep{um2020solver,kochkov2021machine}, while iterative refinement progressively corrects predicted states~\citep{lippe2023pde}. 
Structure-aware approaches impose properties such as flow-map composition or temporally consistent latent transitions~\citep{chen2023deep,nayak2025temporally}. 
Collectively, these approaches address rollout errors through iterative correction or explicit structural constraints.
However, they do not generally characterize how discrepancies induced by different rollouts propagate through the solver's internal evolution, leaving the dependence on temporal origin only indirectly constrained.

We study autoregressive error accumulation from the perspective of consistency across temporal origins. 
For a well-posed deterministic PDE flow with fixed governing quantities, states sampled at different times along the same physical trajectory should, under identical governing conditions, evolve to the same state at a shared future time. 
We refer to this shared-target consequence of deterministic evolution as \emph{cross-origin consistency}.
In a learned solver, however, different temporal origins imply different rollout depths and hence different error histories, which can lead to origin-dependent predictions at the same target time.
Standard prediction losses implicitly encourage cross-origin consistency by aligning physical outputs with the ground-truth trajectory, but do not explicitly couple the internal representations produced by different rollout origins at a shared target time.
Since neural PDE solvers typically realize learned evolution through latent variables or intermediate features~\citep{wu2022lepde,kontolati2024learning,wang2024latent}, cross-origin consistency in this representation space remains underconstrained.

Motivated by this observation, we propose \textbf{A}lign \textbf{L}atent \textbf{R}ollout (\textbf{ALR}), a latent-space regularization approach that aligns latent states produced by different rollout origins at a shared target time. 
A second-order Taylor analysis characterizes how a shared-target latent discrepancy propagates through the learned evolution. This motivates two complementary objectives: zeroth-order latent state consistency reduces the mismatch between shared-target representations, while first-order latent dynamics consistency aligns their local responses to perturbations.
The latter addresses differences in local evolution that can persist when state alignment is imperfect. We estimate this objective using stochastic Jacobian–vector products, avoiding explicit construction of the full Jacobians.
ALR operates on suitable internal representations and their induced local evolution, allowing it to be integrated into a general neural PDE solver without modifying its backbone architecture.
Our main contributions are summarized as follows:
\begin{itemize}[leftmargin=1.2em, itemsep=2pt, topsep=2pt, parsep=0pt]

\item We formulate \emph{cross-origin consistency} as a principle for autoregressive PDE prediction, linking error accumulation to shared-target discrepancies across rollout origins.

\item We propose ALR, a latent-space regularization approach motivated by a second-order Taylor analysis of latent discrepancy propagation. ALR introduces zeroth-order latent state consistency and first-order latent dynamics consistency to align shared-target representations and their local evolution responses to common perturbations, respectively.

\item We evaluate ALR across multiple time-dependent PDE benchmarks and neural-operator backbones. Quantitative and qualitative results demonstrate improved prediction accuracy and reduced latent state and dynamics discrepancies across rollout origins.

\end{itemize}

\vspace{-10pt}
\section{Related Work}

\textbf{Neural Operators for Time-Dependent PDEs.}
Neural operators learn function-space mappings and are widely used for time-dependent PDEs~\citep{kovachki2023neural}.
Foundational formulations parameterize solution operators via branch-trunk decompositions or learned kernels~\citep{lu2021learning,li2020neural}.
Spectral architectures capture global interactions in Fourier or wavelet bases, while factorized and multiscale designs improve efficiency and resolution handling~\citep{li2020fno,gupta2021multiwavelet,tran2023factorized,rahman2022uno,wen2022ufno}.
Geometry-aware and convolutional formulations extend operator learning to irregular domains and discretizations~\citep{li2023geofno,raonic2023cno,serrano2023operator}.
Attention-based and latent-space architectures model heterogeneous interactions and compact spatiotemporal dynamics~\citep{cao2021choose,li2023scalableTransformer,hao2023gnot,wu2023lsm,wu2024transolver,wang2024latent}.

\textbf{Autoregressive Prediction and Rollout Error Mitigation.}
When neural PDE solvers are applied recursively, prediction errors are fed back as inputs and may accumulate over the rollout.
Data-centric strategies mitigate this effect by mixing ground-truth and predicted inputs, perturbing training states, or optimizing multistep rollouts~\citep{ross2011reduction,bengio2015scheduled,pfaff2020learning,brandstetter2022message}.
Solver-in-the-loop approaches differentiate through interactions between learned components and numerical solvers~\citep{um2020solver,kochkov2021machine}.
Other methods improve rollout behavior by controlling unstable operations, matching invariant statistics, or constraining spectral evolution~\citep{mccabe2023stability,schiff2024dyslim,li2026sgno}.
Iterative refinement, trajectory-level generative modeling, and continuous latent evolution provide alternatives to direct autoregressive prediction~\citep{lippe2023pde,hu2025wavelet,zhang2026autoregression}.
Despite their effectiveness, these approaches generally do not characterize how discrepancies across rollout histories propagate through learned evolution, leaving temporal-origin sensitivity largely implicit.

\textbf{Latent Dynamics Regularization.}
Latent-dynamics models encode high-dimensional observations into compact representations and learn continuous-time or approximately linear evolution in latent space~\citep{takeishi2017learning,lusch2018deep,rubanova2019latent}.
Structure-preserving variants incorporate Hamiltonian, Lagrangian, symplectic, or thermodynamic constraints into learned dynamics~\citep{greydanus2019hamiltonian,cranmer2020lagrangian,zhong2019symplectic,he2025thermodynamically}.
For PDE systems, reduced-order and latent-operator models evolve compact representations to capture spatiotemporal dynamics efficiently~\citep{wu2022lepde,yin2022continuous,regazzoni2024learning,tiwari2025latent}.
Consistency-based approaches impose forward--backward agreement or temporal compatibility across multistep latent evolution~\citep{azencot2020forecasting,liu2026semigroup}.
A recent work aligns latent forecasts initialized at different times under Koopman-linear dynamics, primarily at the state level~\citep{nayak2025temporally}.
In this context, we consider consistency over suitable intermediate representations of neural PDE solvers at both shared-target states and local first-order evolution responses, without assuming a particular latent evolution structure.

\section{Method}
We introduce Align Latent Rollout (ALR), which regularizes latent representations obtained from different temporal origins at a shared prediction target. We first formulate cross-origin consistency in deterministic PDE evolution, then analyze the propagation of latent discrepancies to motivate the latent state and dynamics consistency objectives.

\begin{figure*}[!t]
    \centering
    \includegraphics[width=\textwidth]{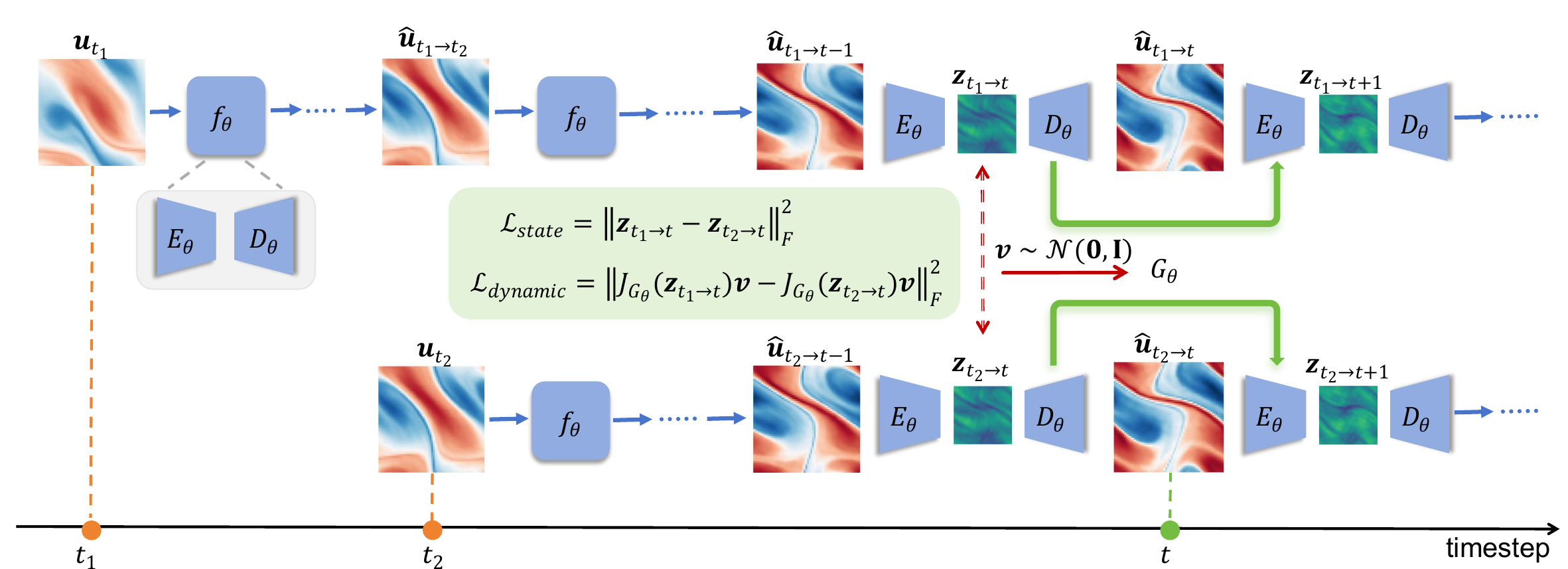}
    \vspace{-15pt}
    \caption{\textbf{Overview of Align Latent Rollout (ALR).}
    Starting from temporal origins along the same physical trajectory, ALR aligns the latent states reached at a shared target time and further regularizes their local evolution through zeroth-order state consistency and first-order dynamics consistency.}
    \label{fig:alr_overview}
    \vspace{-15pt}
\end{figure*}

\subsection{Problem Setting}
\label{sec:problem_setting}
\textbf{PDE Formulation.}
We consider time-dependent parametric PDEs defined on a spatial domain $\Omega\subset\mathbb{R}^{d}$ over a time interval $[0,T]$.
Let $\boldsymbol{u}:\Omega\times[0,T]\rightarrow\mathbb{R}^{d_u}$ denote the solution field, and let $\boldsymbol{\mu}$ denote the governing configuration, including the PDE coefficients, external forcing, and boundary data.
For a fixed governing configuration $\boldsymbol{\mu}$, we assume the PDE to be deterministic and well-posed, such that each admissible initial condition $\boldsymbol{u}_0$ uniquely determines a forward solution trajectory:
\begin{equation}
\begin{aligned}
\mathcal{N}[\boldsymbol{u};\boldsymbol{\mu}](\boldsymbol{x},\tau)
&=0,
&& (\boldsymbol{x},\tau)\in\Omega\times(0,T],\\
\mathcal{B}[\boldsymbol{u};\boldsymbol{\mu}](\boldsymbol{x},\tau)
&=0,
&& (\boldsymbol{x},\tau)\in\partial\Omega\times[0,T],\\
\boldsymbol{u}(\boldsymbol{x},0)
&=\boldsymbol{u}_0(\boldsymbol{x}),
&& \boldsymbol{x}\in\Omega.
\end{aligned}
\label{eq:pde}
\end{equation}
where $\mathcal{N}$ and $\mathcal{B}$ denote the governing differential and boundary operators, respectively.

\textbf{Neural Operator Formulation.} 
Given a uniform temporal step $\Delta\tau$, we denote the discrete state at index $t$ by $\boldsymbol{u}_t:=\boldsymbol{u}(\cdot,t\Delta\tau)$, for $t=0,\ldots,N$.
For input and output window lengths $T_{\mathrm{in}}$ and $T_{\mathrm{out}}$, we define input history ending at $t$ as
$\boldsymbol{U}_t=(\boldsymbol{u}_{t-T_{\mathrm{in}}+1},\ldots,\boldsymbol{u}_t)$, and future window beginning at $t+1$ as
$\boldsymbol{V}_{t+1}= (\boldsymbol{u}_{t+1},\ldots,\boldsymbol{u}_{t+T_{\mathrm{out}}})$.
Under governing configuration $\boldsymbol{\mu}$, a neural operator $f_\theta$ predicts the future window from the input history:
\begin{equation}
\begin{aligned}
\hat{\boldsymbol{V}}_{t+1}
=
f_\theta(\boldsymbol{U}_t;\boldsymbol{\mu})
\end{aligned}
\label{eq:neural_operator}
\end{equation}

We omit the explicit dependence on $\boldsymbol{\mu}$ below when the governing configuration is fixed or clear from context.
For notational clarity, we present the subsequent analysis with
$T_{\mathrm{in}}=T_{\mathrm{out}}=1$, while the general formulation remains applicable to arbitrary temporal windows.
The learned operator then defines the one-step prediction
$\hat{\boldsymbol{u}}_{t+1}=f_\theta(\boldsymbol{u}_t)$.

\textbf{Autoregressive Rollout.}
At inference, multi-step predictions are generated autoregressively by repeatedly applying $f_\theta$ and feeding each prediction back.
Let $t_i$ denote the rollout origin and $h\in\mathbb{Z}_{>0}$ the rollout horizon.
Starting from $\boldsymbol{u}_{t_i}$, the $h$-step rollout prediction is defined as:
\begin{equation}
\hat{\boldsymbol{u}}_{t_i\rightarrow t_i+h}
=
f_\theta^{(h)}(\boldsymbol{u}_{t_i}),
\qquad
f_\theta^{(h)}
:=
\underbrace{
f_\theta\circ f_\theta\circ\cdots\circ f_\theta
}_{h\ \mathrm{times}}.
\label{eq:autoregressive_rollout}
\end{equation}
Here, the subscript $t_i\rightarrow t_i+h$ records the rollout origin $t_i$ and target time $t_i+h$, while $f_\theta^{(h)}$ denotes the $h$-fold composition of $f_\theta$.
More generally, for $t_i<t_j$,
$\hat{\boldsymbol{u}}_{t_i\rightarrow t_j}
=
f_\theta^{(t_j-t_i)}(\boldsymbol{u}_{t_i})$.

Although multi-step inference is autoregressive, $f_\theta$ may be trained with different prediction objectives.
Under one-step training, each prediction is conditioned on the ground-truth state at the preceding time step:
$\hat{\boldsymbol{u}}_{t-1\rightarrow t}
=
f_\theta^{(1)}(\boldsymbol{u}_{t-1})
=
f_\theta(\boldsymbol{u}_{t-1})$.
In contrast, under unrolled training, the model may start from an earlier ground-truth state $\boldsymbol{u}_{t-k}$ and recursively feed back its predictions, yielding
$\hat{\boldsymbol{u}}_{t-k\rightarrow t}
=
f_\theta^{(k)}(\boldsymbol{u}_{t-k})$,
where $k\in\mathbb{Z}_{>0}$.

Consequently, the same target $\boldsymbol{u}_t$ can be reached from different temporal origins, e.g., by a one-step prediction from $\boldsymbol{u}_{t-1}$ or a $k$-step rollout from $\boldsymbol{u}_{t-k}$.
Learned rollouts, however, may accumulate path-dependent approximation errors, producing different predictions at the same target time.
This shared-target discrepancy motivates our study of consistency across temporal origins.

\subsection{Cross-Origin Consistency in Deterministic PDE Evolution}
\label{sec:path_consistency}

\textbf{Deterministic PDE Flow and Cross-Origin Consistency.}
For a fixed governing configuration, the deterministic well-posedness assumption above defines a unique forward evolution from each admissible initial state.
Exact states sampled at different times along the same trajectory therefore remain compatible under forward evolution to a common target time.

Consider three states $\boldsymbol{u}_{t_1}$, $\boldsymbol{u}_{t_2}$, and $\boldsymbol{u}_t$ along the same solution trajectory, where $t_1<t_2<t$, with $t_1,t_2$ as temporal origins and $t$ as the shared target time.
Let $\Phi_{t_1\rightarrow t_2}$, $\Phi_{t_2\rightarrow t}$, and
$\Phi_{t_1\rightarrow t}$ denote the corresponding exact evolution operators, such that
$\boldsymbol{u}_{t_2} = \Phi_{t_1\rightarrow t_2}(\boldsymbol{u}_{t_1})$.
By the composition of deterministic PDE evolution, advancing from $t_1$
to $t$ directly is equivalent to evolving through the intermediate time $t_2$:
$
\Phi_{t_1\rightarrow t}
=
\Phi_{t_2\rightarrow t}
\circ
\Phi_{t_1\rightarrow t_2}$.
Therefore,
\begin{equation}
\begin{aligned}
\Phi_{t_1\rightarrow t}(\boldsymbol{u}_{t_1})
=
\Phi_{t_2\rightarrow t}
\left(
\Phi_{t_1\rightarrow t_2}(\boldsymbol{u}_{t_1})
\right) 
=
\Phi_{t_2\rightarrow t}(\boldsymbol{u}_{t_2})
=
\boldsymbol{u}_t
\end{aligned}
\label{eq:physical_path_consistency}
\end{equation}
We refer to this shared-target consequence as \emph{cross-origin consistency}: different temporal origins along the same trajectory reach the same state at a common target time.

\textbf{Cross-Origin Inconsistency in Learned Neural PDE Solvers.}
A learned operator $f_\theta$ approximates the exact evolution but is not explicitly constrained to preserve cross-origin consistency under recursive application.
Different temporal origins imply different rollout depths, along which local approximation errors may propagate differently.

Let $\hat{\boldsymbol{u}}_{t_1\rightarrow t}$ and $\hat{\boldsymbol{u}}_{t_2\rightarrow t}$ denote predictions at the shared target time $t$ from origins $t_1<t_2<t$.
Although both predictions correspond to the same target
$\boldsymbol{u}_t$, they may differ:
$\hat{\boldsymbol{u}}_{t_1\rightarrow t}
\neq
\hat{\boldsymbol{u}}_{t_2\rightarrow t}$.

A standard prediction loss
$\mathcal{L}_{\mathrm{pred}} = \ell(\hat{\boldsymbol{u}}_{t_i\rightarrow t},\boldsymbol{u}_t)$ encourages agreement with the ground-truth trajectory at the supervised physical prediction steps.
However, it does not directly couple the internal representations reached from different origins, which can remain different even when their decoded predictions are close.
As the learned evolution is realized through these representations, origin-dependent latent discrepancies may persist and affect their subsequent evolution.

\textbf{Latent Representation of Learned Evolution.}
Learned representations organize task-relevant information for prediction~\citep{bengio2013representation} and provide internal coordinates for modeling nonlinear dynamics~\citep{lusch2018deep}.
Neural PDE solvers similarly encode physical fields into latent representations before decoding predictions~\citep{wu2022lepde,wu2023lsm,kontolati2024learning,wang2024latent}.
These studies suggest that latent representations can serve as internal coordinates through which neural PDE solvers realize their learned temporal evolution.

We designate a selected intermediate representation as the latent state for analysis.
Specifically, we partition $f_\theta$ at this layer as
$f_\theta=D_\theta\circ E_\theta$, where $E_\theta$ encodes the physical input to the latent representation and $D_\theta$ decodes it to the physical prediction.
This functional decomposition does not require an explicit encoder--decoder architecture and applies whenever a suitable intermediate representation is available.
For a rollout originating at $t_i$, the prediction step reaching time $t$ is represented as
$\hat{\boldsymbol{u}}_{t_i\rightarrow t-1} \xrightarrow{E_\theta} \boldsymbol{z}_{t_i\rightarrow t} \xrightarrow{D_\theta} \hat{\boldsymbol{u}}_{t_i\rightarrow t}$,
where $\boldsymbol{z}_{t_i\rightarrow t}$ is the selected latent representation associated with target time $t$ along the rollout from the origin $t_i$.
Accordingly, we index the latent state by its prediction target $t$, although it is computed by encoding the rollout input at $t-1$.

We therefore compare representations associated with the same prediction target across different rollout histories.
For two rollout origins $t_1<t_2<t$, we seek: 
\begin{equation}
\boldsymbol{z}_{t_1\rightarrow t}
\approx
\boldsymbol{z}_{t_2\rightarrow t}
\label{eq:latent_path_consistency}
\end{equation}
Both representations are extracted from the same layer of the shared operator, and their alignment complements physical-space supervision by constraining origin-dependent latent variation.
Since different rollout histories can still induce shared-target latent discrepancies, we next analyze how these discrepancies propagate under the learned evolution.


\subsection{Align Latent Rollout}
\label{sec:align latent rollout}
Building on the latent-space view above, ALR regularizes cross-origin discrepancies in the internal representations through which the solver realizes its learned evolution. We motivate the regularization objectives by analyzing how such discrepancies propagate under the induced latent evolution.

\textbf{Induced Latent Evolution.}
For a rollout originating at $t_i$, one latent evolution step decodes the latent state into a physical prediction and then re-encodes that prediction: 
$\boldsymbol{z}_{t_i\rightarrow t}
\xrightarrow{D_\theta}
\hat{\boldsymbol{u}}_{t_i\rightarrow t}
\xrightarrow{E_\theta}
\boldsymbol{z}_{t_i\rightarrow t+1}$.
While $f_\theta=D_\theta\circ E_\theta$ maps a physical input to its next prediction, composing the same mappings in the opposite order defines the induced latent evolution map
$G_\theta:=E_\theta\circ D_\theta$, which maps latent representations across consecutive prediction targets:
\begin{equation}
\boldsymbol{z}_{t_i\rightarrow t+1}
=
E_\theta\left(
D_\theta(\boldsymbol{z}_{t_i\rightarrow t})
\right)
=
G_\theta(\boldsymbol{z}_{t_i\rightarrow t}).
\label{eq:latent_evolution}
\end{equation}

\textbf{Taylor Analysis of Latent Discrepancy Propagation.}
Consider two rollouts originating at $t_1<t_2<t$ from states on the same physical trajectory and propagated to the shared target time $t$.
Let their shared-target representations be
$\boldsymbol{z}_{t_1\rightarrow t}$ and $\boldsymbol{z}_{t_2\rightarrow t}$, and define the cross-origin latent discrepancy as
$\Delta\boldsymbol{z}_t
:=
\boldsymbol{z}_{t_1\rightarrow t}
-
\boldsymbol{z}_{t_2\rightarrow t}$.
Following Eq.~\ref{eq:latent_path_consistency}, we seek
$\Delta\boldsymbol{z}_t\approx\boldsymbol{0}$.
In practice, this alignment may not be exact, leaving a residual cross-origin discrepancy.
We therefore analyze how $\Delta\boldsymbol{z}_t$ propagates through the induced latent evolution $G_\theta$ to the next step:
\begin{equation}
\begin{aligned}
\Delta\boldsymbol{z}_{t+1}
=
\boldsymbol{z}_{t_1\rightarrow t+1}
-
\boldsymbol{z}_{t_2\rightarrow t+1} 
=
G_\theta(\boldsymbol{z}_{t_1\rightarrow t})
-
G_\theta(\boldsymbol{z}_{t_2\rightarrow t}) 
=
G_\theta\left(
\boldsymbol{z}_{t_2\rightarrow t}
+
\Delta\boldsymbol{z}_t
\right)
-
G_\theta(\boldsymbol{z}_{t_2\rightarrow t})
\end{aligned}
\label{eq:latent_discrepancy_evolution}
\end{equation}

Assuming $G_\theta$ is locally three times continuously differentiable, a second-order Taylor expansion around $\boldsymbol{z}_{t_2\rightarrow t}$ gives:
\begin{equation}
\begin{aligned}
\Delta\boldsymbol{z}_{t+1}
=
J_{G_\theta}(\boldsymbol{z}_{t_2\rightarrow t})
\Delta\boldsymbol{z}_t
+
\frac{1}{2}
H_{G_\theta}(\boldsymbol{z}_{t_2\rightarrow t})
\left[
\Delta\boldsymbol{z}_t,
\Delta\boldsymbol{z}_t
\right]
+
O\left(\|\Delta\boldsymbol{z}_t\|^3\right)
\end{aligned}
\label{eq:latent_taylor}
\end{equation}
where $J_{G_\theta}(\boldsymbol z)$ denotes the Jacobian of $G_\theta$, and $H_{G_\theta}(\boldsymbol z)[\Delta\boldsymbol z,\Delta\boldsymbol z]$ its second directional derivative along $\Delta\boldsymbol z$.
The first term describes local linearized discrepancy propagation, while the second represents the leading correction induced by the local curvature of $G_\theta$.

To express this curvature correction through variation in the local linearized evolution, we expand the Jacobian at $\boldsymbol{z}_{t_1\rightarrow t}$ around $\boldsymbol{z}_{t_2\rightarrow t}$:
\begin{equation}
\begin{aligned}
J_{G_\theta}(\boldsymbol{z}_{t_1\rightarrow t})
=
J_{G_\theta}
\left(
\boldsymbol{z}_{t_2\rightarrow t}
+
\Delta\boldsymbol{z}_t
\right) 
=
J_{G_\theta}(\boldsymbol{z}_{t_2\rightarrow t})
+
H_{G_\theta}(\boldsymbol{z}_{t_2\rightarrow t})
\left[
\Delta\boldsymbol{z}_t
\right]
+
O\left(\|\Delta\boldsymbol{z}_t\|^2\right)
\end{aligned}
\label{eq:jacobian_expansion}
\end{equation}

It follows that:
\begin{equation}
\begin{aligned}
J_{G_\theta}(\boldsymbol{z}_{t_1\rightarrow t})
-
J_{G_\theta}(\boldsymbol{z}_{t_2\rightarrow t})
=
H_{G_\theta}(\boldsymbol{z}_{t_2\rightarrow t})
\left[
\Delta\boldsymbol{z}_t
\right]
+
O\left(\|\Delta\boldsymbol{z}_t\|^2\right)
\end{aligned}
\label{eq:jacobian_variation}
\end{equation}

Applying both sides of Eq.~\ref{eq:jacobian_variation} to
$\Delta\boldsymbol{z}_t$ yields:
\begin{equation}
\begin{aligned}
\left(
J_{G_\theta}(\boldsymbol{z}_{t_1\rightarrow t})
-
J_{G_\theta}(\boldsymbol{z}_{t_2\rightarrow t})
\right)
\Delta\boldsymbol{z}_t
=
H_{G_\theta}(\boldsymbol{z}_{t_2\rightarrow t})
\left[
\Delta\boldsymbol{z}_t,
\Delta\boldsymbol{z}_t
\right]
+
O\left(\|\Delta\boldsymbol{z}_t\|^3\right)
\end{aligned}
\label{eq:jacobian_variation_direction}
\end{equation}

Substituting this relation into Eq.~\ref{eq:latent_taylor} gives
\begin{equation}
\begin{aligned}
\Delta\boldsymbol{z}_{t+1}
=
J_{G_\theta}(\boldsymbol{z}_{t_2\rightarrow t})
\Delta\boldsymbol{z}_t
+
\frac{1}{2}
\left(
J_{G_\theta}(\boldsymbol{z}_{t_1\rightarrow t})
-
J_{G_\theta}(\boldsymbol{z}_{t_2\rightarrow t})
\right)
\Delta\boldsymbol{z}_t
+
O\left(\|\Delta\boldsymbol{z}_t\|^3\right)
\end{aligned}
\label{eq:cross_path_evolution}
\end{equation}
In Eq.~\ref{eq:cross_path_evolution}, the first term propagates the existing mismatch through the local Jacobian at $\boldsymbol{z}_{t_2\rightarrow t}$.
The second term captures the leading correction induced by variation of the local Jacobian across the two shared-target representations.
This decomposition motivates two complementary objectives: zeroth-order state consistency reduces the latent mismatch itself, while first-order dynamics consistency regularizes cross-origin variation in the local linearized evolution.

\textbf{Zeroth-Order Latent State Consistency.}
To reduce the shared-target latent mismatch in Eq.~\ref{eq:cross_path_evolution}, we encourage
$\Delta\boldsymbol{z}_t\approx\boldsymbol{0}$ by aligning the corresponding latent states:
\begin{equation}
\mathcal{L}_{\mathrm{state}}
=
\left\|
\boldsymbol{z}_{t_1\rightarrow t}
-
\boldsymbol{z}_{t_2\rightarrow t}
\right\|_F^2
\label{eq:state_consistency}
\end{equation}
where $\|\cdot\|_F$ denotes the Frobenius norm.
This objective directly reduces the cross-origin discrepancy between shared-target representations.

\textbf{First-Order Latent Dynamics Consistency.}
The zeroth-order objective aligns shared-target latent states but does not explicitly constrain their subsequent local evolution.
The second term in Eq.~\ref{eq:cross_path_evolution} involves cross-origin variation in the local Jacobian, motivating us to further align the local linearized dynamics by penalizing the full Jacobian discrepancy:
\begin{equation}
\mathcal{L}_{\mathrm{dynamic}}^{\mathrm{full}} 
=  
\left\| J_{G_\theta}\left(\boldsymbol{z}_{t_1\rightarrow t}\right) - J_{G_\theta}\left(\boldsymbol{z}_{t_2\rightarrow t}\right)\right\|_F^2.
\label{eq:dynamic_consistency}
\end{equation}
This objective encourages the two shared-target representations to exhibit consistent first-order responses to latent perturbations.

Explicitly constructing full Jacobians, however, is expensive in high-dimensional latent spaces.
We therefore probe the local evolution along a stochastic direction $\boldsymbol{v}\sim\mathcal{N}(\boldsymbol{0},\mathbf{I})$
using Jacobian--vector products (JVPs), where
$J_{G_\theta}(\boldsymbol{z})\boldsymbol{v}$
represents the local linearized evolution response along $\boldsymbol{v}$.
Using the same probing direction for both representations allows us to compare their responses to a common perturbation.
Since $\mathbb{E}_{\boldsymbol{v}} [\boldsymbol{v}\boldsymbol{v}^{\top}]
= \mathbf{I}$,
we have:
\begin{equation} 
\begin{aligned} 
\mathbb{E}_{\boldsymbol{v}} \left[ \left\| \left( J_{G_\theta}(\boldsymbol{z}_{t_1\rightarrow t}) - J_{G_\theta}(\boldsymbol{z}_{t_2\rightarrow t}) \right) \boldsymbol{v} \right\|_F^2 \right] 
= \left\| J_{G_\theta}(\boldsymbol{z}_{t_1\rightarrow t}) - J_{G_\theta}(\boldsymbol{z}_{t_2\rightarrow t}) \right\|_F^2
\end{aligned} 
\label{eq:jvp_expectation} 
\end{equation}

Accordingly, we optimize the Jacobian discrepancy along a single stochastic probing direction:
\begin{equation}
\mathcal{L}_{\mathrm{dynamic}}
=
\left\|
J_{G_\theta}(\boldsymbol{z}_{t_1\rightarrow t})\boldsymbol{v}
-
J_{G_\theta}(\boldsymbol{z}_{t_2\rightarrow t})\boldsymbol{v}
\right\|_F^2,
\qquad
\boldsymbol{v}\sim\mathcal{N}(\boldsymbol{0},\mathbf{I})
\label{eq:jvp_dynamic_consistency}
\end{equation}
The same probing direction is used for both origins, so the difference between the two JVPs is the action of the Jacobian difference on a common perturbation.
By Eq.~\ref{eq:jvp_expectation}, the resulting loss is an unbiased stochastic estimator of $\mathcal{L}_{\mathrm{dynamic}}^{\mathrm{full}}$.
The probe $\boldsymbol{v}$ is resampled during training, and the required JVPs can be computed efficiently through forward-mode automatic differentiation using \texttt{torch.func.jvp}, without explicitly constructing the Jacobian.

\textbf{Overall Training Objective.}
We retain the original prediction loss $\mathcal{L}_{\mathrm{pred}}$ and augment it with the proposed state and dynamics consistency objectives:
\begin{equation}
\mathcal{L}_{\mathrm{total}}
=
\mathcal{L}_{\mathrm{pred}}
+
\lambda_{\mathrm{state}}\mathcal{L}_{\mathrm{state}}
+
\lambda_{\mathrm{dynamic}}\mathcal{L}_{\mathrm{dynamic}}.
\label{eq:total_loss}
\end{equation}
where $\lambda_{\mathrm{state}}$ and $\lambda_{\mathrm{dynamic}}$ weight the state and dynamics regularization, respectively.

\section{Experiment}

\subsection{Experimental Setup}
We use the FNO~\citep{li2020fno} as the primary backbone and evaluate it on four spatiotemporal PDE benchmarks: Navier--Stokes flows~\citep{li2020fno} with viscosities $\nu=10^{-4}$ and $\nu=10^{-5}$, the Shallow Water equations~\citep{takamoto2022pdebench}, and Kolmogorov Flow~\citep{kochkov2021machine}.
The FNO backbone contains four Fourier blocks, and the output of the second block is used as the latent representation.
We use a learning rate of $5\times10^{-4}$, a weight decay of $10^{-5}$, and a OneCycle learning-rate schedule.
The prediction objective is the relative $\ell_2$ loss, with $\lambda_{\mathrm{state}}=\lambda_{\mathrm{dynamic}}=0.1$.

\textbf{Cross-Origin Consistency Instantiation.}
Under the standard one-step training protocol, we retain the original ground-truth-conditioned prediction objective and construct cross-origin pairs using an auxiliary local autoregressive rollout.
For each training trajectory, we randomly sample an origin $t_i$ and roll out for at most $K$ steps, with $K=6$ by default and truncation near the trajectory end.
Since the $h=1$ local and reference paths share the same ground-truth input and thus yield a trivial pair, consistency is imposed for $h=2,\ldots,K_i$, where $K_i=\min(K,N-t_i)$.
At each shared target $t_i+h$, we align the rollout representation $\boldsymbol{z}_{t_i\rightarrow t_i+h}$ with the ground-truth-conditioned one-step representation $\boldsymbol{z}_{t_i+h-1\rightarrow t_i+h}$ using $\mathcal{L}_{\mathrm{state}}$ and $\mathcal{L}_{\mathrm{dynamic}}$.
The auxiliary rollout is used only for consistency regularization, leaving the original prediction objective unchanged.


\vspace{-8pt}
\subsection{Main Results}
\label{sec:exp_main}

Table~\ref{tab:main_results} reports the results on four spatiotemporal PDE benchmarks.
All models are initialized from a single initial state and autoregressively evaluated for 29, 19, 20, and 20 steps on Navier--Stokes with $\nu=10^{-4}$, Navier--Stokes with $\nu=10^{-5}$, Shallow Water, and Kolmogorov Flow, respectively.
Compared with the FNO baseline, FNO+ALR consistently reduces the relative $\ell_2$ error across all benchmarks, with relative improvements of $29.25\%$, $14.54\%$, $28.95\%$, and $31.77\%$, respectively.
These results demonstrate the effectiveness of cross-origin consistency regularization across diverse spatiotemporal dynamics.

\setlength{\textfloatsep}{4pt}
\setlength{\intextsep}{4pt}
\begin{table}[!htbp]
\centering
\caption{Results on four spatiotemporal PDE benchmarks.
Unless otherwise specified, prediction errors hereafter are reported as mean $\pm$ standard deviation of relative $\ell_2$ error ($\downarrow$).
FNO is the one-step baseline, FNO+ALR augments it with cross-orgin consistency regularization, and FNO (Unroll) is an unrolled-training reference.
Percentages denote relative error reductions over the FNO baseline.}
\label{tab:main_results}

\small

\begin{tabular}{@{}lcccc@{}}
\toprule
Method
& NS ($\nu=10^{-4}$)
& NS ($\nu=10^{-5}$)
& Shallow Water
& Kolmogorov Flow \\
\midrule

FNO
& $0.0947 \pm 0.0012$
& $0.1004 \pm 0.0015$
& $0.0038 \pm 0.0002$
& $0.4095 \pm 0.0164$ \\
\midrule

FNO+ALR (Ours)
& \begin{tabular}[c]{@{}c@{}}
  $\mathbf{0.0670} \pm 0.0004$ \\
  {$\downarrow\,29.25\%$}
  \end{tabular}
& \begin{tabular}[c]{@{}c@{}}
  $\mathbf{0.0858} \pm 0.0003$ \\
  {$\downarrow\,14.54\%$}
  \end{tabular}
& \begin{tabular}[c]{@{}c@{}}
  $\mathbf{0.0027} \pm 0.0001$ \\
  { $\downarrow\,28.95\%$}
  \end{tabular}
& \begin{tabular}[c]{@{}c@{}}
  $\mathbf{0.2794} \pm 0.0089$ \\
  {$\downarrow\,31.77\%$}
  \end{tabular} \\

\midrule

FNO (Unroll)
& $0.2184 \pm 0.0035$
& $0.1682 \pm 0.0021$
& $0.0032 \pm 0.0002$
& $0.3445 \pm 0.0103$ \\

\bottomrule
\end{tabular}
\end{table}


\subsection{Ablation Study}

\begin{table}[!htbp]
\vspace{-7pt}
\centering

\begin{minipage}[t]{0.48\textwidth}
\vspace{0pt}
\centering

\captionof{table}{Ablation of the zeroth-order state and first-order dynamics consistency objectives on FNO. Relative $\ell_2$ errors ($\downarrow$) are reported on Navier--Stokes (NS) with $\nu=10^{-5}$ and Shallow Water (SW).}
\label{tab:loss_ablation}

\small
\setlength{\tabcolsep}{4pt}

\begin{tabular}{cccc}
\toprule
$\mathcal{L}_{\mathrm{state}}$
& $\mathcal{L}_{\mathrm{dynamic}}$
& NS ($\nu=10^{-5}$)
& SW \\
\midrule
$\times$       & $\times$     & 0.1004 & 0.0038 \\
\checkmark     & $\times$     & 0.0961 & 0.0034 \\
$\times$       & \checkmark   & 0.0887 & 0.0029 \\
\checkmark     & \checkmark   & \textbf{0.0858} & \textbf{0.0027} \\
\bottomrule
\end{tabular}

\end{minipage}
\hfill
\begin{minipage}[t]{0.48\textwidth}
\vspace{0pt}
\centering

\captionof{table}{Effect of the consistency-window length $K$ on Navier--Stokes (NS) with $\nu=10^{-5}$ and Shallow Water (SW). Relative $\ell_2$ errors ($\downarrow$) are reported.}
\label{tab:k_ablation}

\vspace{4pt}
\small
\setlength{\tabcolsep}{4pt}
\begin{tabular}{ccc}
\toprule
$K$
& NS ($\nu=10^{-5}$)
& SW \\
\midrule
$-$  & 0.1004 & 0.0038 \\
4    & 0.0893 & 0.0028 \\
6    & 0.0858 & \textbf{0.0027} \\
8    & \textbf{0.0851} & \textbf{0.0027} \\
10   & 0.0856 & \textbf{0.0027} \\
\bottomrule
\end{tabular}

\end{minipage}

\end{table}

\textbf{Quantitative Analysis of Consistency Objectives.}
Table~\ref{tab:loss_ablation} evaluates the two consistency objectives individually and jointly.
On Navier-- Stokes with $\nu=10^{-5}$, $\mathcal{L}_{\mathrm{state}}$ reduces the error from $0.1004$ to $0.0961$, while $\mathcal{L}_{\mathrm{dynamic}}$ achieves a larger reduction to $0.0887$; the same trend is observed on Shallow Water, where the error decreases from $0.0038$ to $0.0034$ and $0.0029$, respectively.
The larger gain from $\mathcal{L}_{\mathrm{dynamic}}$ highlights the importance of explicitly regularizing local latent evolution beyond state alignment alone.
Combining both objectives yields the lowest errors of $0.0858$ on Navier--Stokes and $0.0027$ on Shallow Water, demonstrating the complementary roles of zeroth-order state and first-order dynamics consistency.


\begin{figure}[!t] 
    \vspace{-10pt}
    \centering 
    \includegraphics[width=\linewidth]{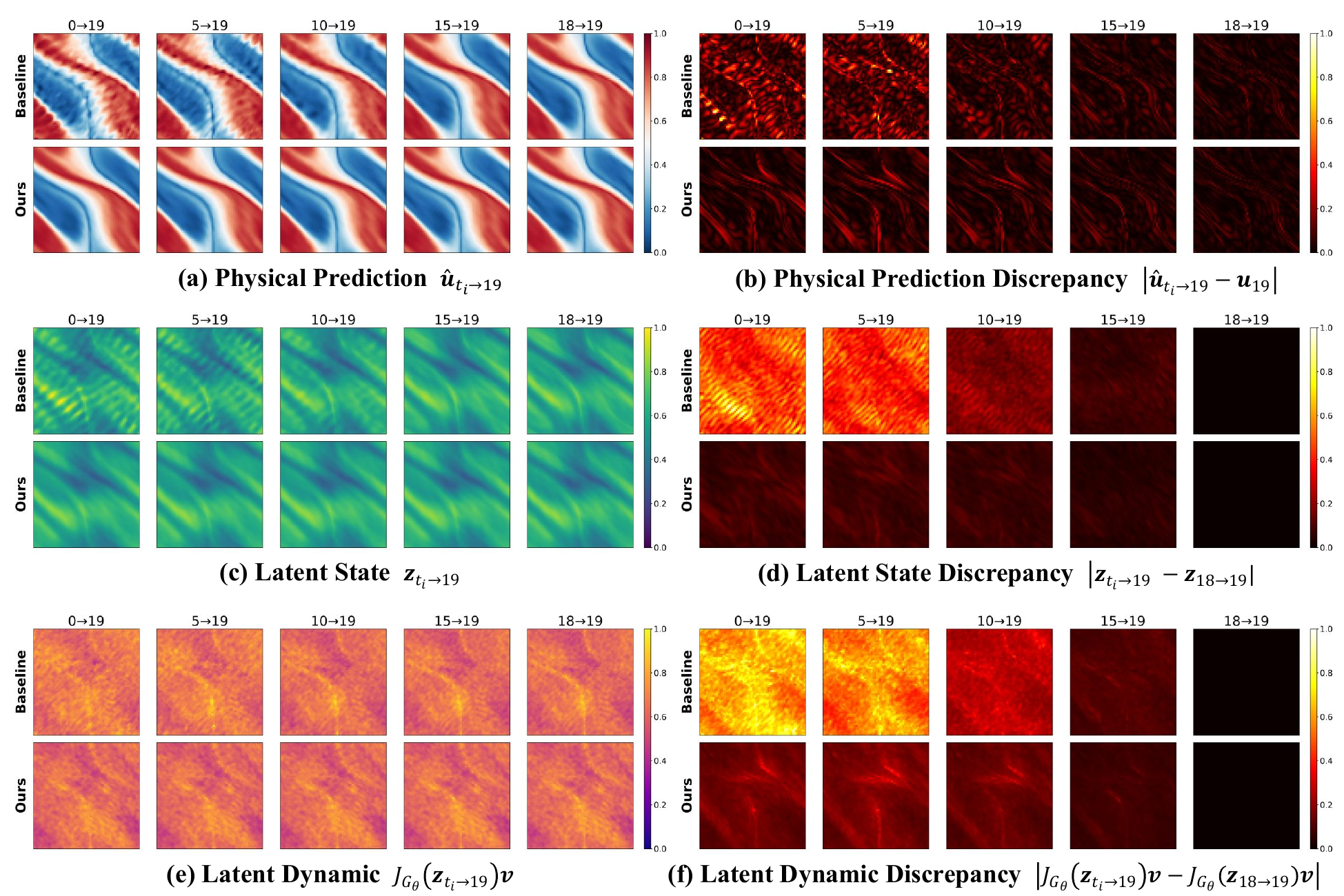} 
    \caption{Qualitative analysis of cross-path consistency on the Navier--Stokes benchmark ($\nu=10^{-5}$).
        Each column corresponds to a rollout origin $t_i\in\{0,5,10,15,18\}$, with all paths propagated to the shared target time $t=19$.
        The top and bottom rows of each panel show the baseline FNO and FNO+ALR, respectively.
        (a) Physical predictions at $t=19$.
        (b) Physical prediction errors relative to the ground truth.
        (c) Latent states at $t=19$.
        (d) Latent-state discrepancies relative to the ground-truth-conditioned one-step path $18\rightarrow19$.
        (e) Local latent dynamics represented by $J_{G_\theta}(\boldsymbol{z}_{t_i\rightarrow 19})\boldsymbol{v}$, using the
        same probing direction $\boldsymbol{v}$ for all paths.
        (f) Corresponding latent-dynamics discrepancies relative to the same reference path.
        The rightmost columns in (d) and (f) are zero by construction because the reference path is compared with itself.
    }
    \label{fig:qualitative_consistency}
    \vspace{5pt}
\end{figure}

\textbf{Qualitative Analysis of Cross-Origin Consistency.}
Figure~\ref{fig:qualitative_consistency} compares rollouts from temporal origins
$t_i\in\{0,5,10,15,18\}$ to the shared target time $t=19$, with $t_i=18$ used as the one-step reference for latent-space discrepancies.
Figures~\ref{fig:qualitative_consistency}(a) and (b) show the physical predictions and their errors relative to the ground truth.
For the baseline FNO, earlier origins produce increasingly distorted predictions and larger errors, whereas FNO+ALR maintains substantially more consistent predictions across rollout depths.

Figures~\ref{fig:qualitative_consistency}(c) and (d) further reveal the corresponding behavior in latent space.
The baseline exhibits pronounced cross-origin variation in $\boldsymbol{z}_{t_i\rightarrow 19}$, particularly for earlier origins, while FNO+ALR markedly reduces the discrepancies relative to $\boldsymbol{z}_{18\rightarrow 19}$.
This behavior is consistent with $\mathcal{L}_{\mathrm{state}}$, which directly suppresses the shared-target latent discrepancy $\Delta\boldsymbol{z}_t$.
Figures~\ref{fig:qualitative_consistency}(e) and (f) visualize the local evolution responses
$J_{G_\theta}(\boldsymbol{z}_{t_i\rightarrow 19})\boldsymbol{v}$ under the same probing direction $\boldsymbol{v}$.
The strong origin-dependent variation observed in the baseline is substantially reduced by ALR, supporting the role of $\mathcal{L}_{\mathrm{dynamic}}$ in aligning the latent evolution across origins.
Together, (c)--(f) provide qualitative support for our discrepancy-propagation analysis: ALR regularizes both the shared-target latent states and their local evolution, which is accompanied by the improved physical predictions in (a) and (b).


\textbf{Effect of Consistency-Window Length $K$.}
The consistency-window length $K$ controls the rollout depth used to construct cross-origin consistency pairs.
As shown in Table~\ref{tab:k_ablation}, increasing $K$ from $4$ to $6$ reduces the relative $\ell_2$ error from $0.0893$ to $0.0858$ on Navier--Stokes and from $0.0028$ to $0.0027$ on Shallow Water.
Further increasing $K$ provides only marginal gains: Navier--Stokes reaches $0.0851$ at $K=8$ and $0.0856$ at $K=10$, while Shallow Water remains at $0.0027$.
We therefore use $K=6$ by default, which captures most of the improvement with a moderate rollout depth.
Although larger $K$ increases training cost through longer auxiliary rollouts and additional JVP evaluations, inference remains unchanged from the FNO baseline.


\subsection{Generalization Across Backbones}
We further evaluate ALR on two additional backbones, F-FNO and LSM.
We follow the experimental settings in Section~\ref{sec:exp_main} and retain the same ALR hyperparameters, while selecting an appropriate intermediate representation as the latent state for each architecture.
As shown in Table~\ref{tab:backbone_generalization}, ALR consistently improves both backbones on the two PDE benchmarks.
For F-FNO, the relative $\ell_2$ error decreases from $0.2904$ to $0.2126$ on Navier--Stokes and from $0.0073$ to $0.0063$ on Shallow Water; for LSM, it decreases from $0.2812$ to $0.2188$ and from $0.0021$ to $0.0019$, respectively.
Notably, LSM explicitly models an encoder--latent processor--decoder pipeline, whereas F-FNO provides the latent state through an intermediate operator representation.
The consistent improvements across these distinct architectures indicate that ALR is not tied to a specific backbone or latent-state realization.

\begin{table}[!htbp]
\centering
\caption{Generalization of ALR across neural operator backbones on Navier--Stokes (NS) and Shallow Water (SW), measured by relative $\ell_2$ error ($\downarrow$). Percentages denote relative error reductions over the corresponding backbone.}
\label{tab:backbone_generalization}

\small
\setlength{\tabcolsep}{5pt}
\renewcommand{\arraystretch}{0.95}

\begin{tabular}{@{}lcc c lcc@{}}
\toprule
Method
& NS ($\nu=10^{-4}$)
& SW
&
& Method
& NS ($\nu=10^{-4}$)
& SW \\
\midrule

F-FNO
& 0.2904
& 0.0073
& $\vert$
& LSM
& 0.2812
& 0.0021 \\

F-FNO+ALR
& \textbf{0.2126}\,{\scriptsize($\downarrow 26.79\%$)}
& \textbf{0.0063}\,{\scriptsize($\downarrow 13.70\%$)}
& $\vert$
& LSM+ALR
& \textbf{0.2188}\,{\scriptsize($\downarrow 22.19\%$)}
& \textbf{0.0019}\,{\scriptsize($\downarrow 9.52\%$)} \\

\bottomrule
\end{tabular}

\vspace{4pt}
\end{table}















\subsection{Performance Across Rollout Lengths}
\label{sec:rollout_length}

\begin{figure*}[!htbp]
    \centering
    \includegraphics[width=0.95\textwidth]{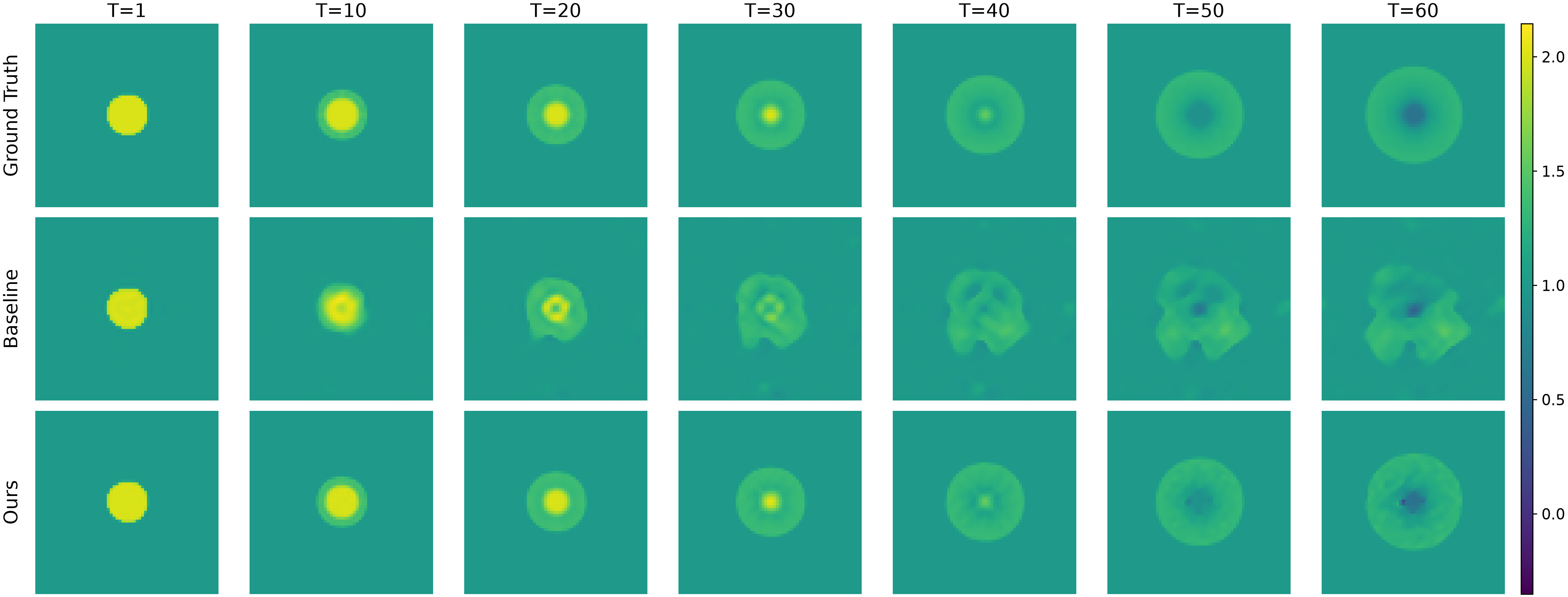}
    \vspace{-6pt}
    \caption{Qualitative rollout prediction results on Shallow Water under the rollout-length-$60$ training setting.
    Ground truth, FNO, and FNO+ALR are shown from top to bottom at selected rollout steps
    $T\in\{1,10,20,30,40,50,60\}$.}
    \label{fig:shallow_water_long_rollout}
\end{figure*}

\begin{wraptable}{r}{0.45\textwidth}
\vspace{-14pt}
\centering
\caption{Autoregressive rollout performance at increasing horizons on the Shallow Water benchmark. Results are reported using relative $\ell_2$ error ($\downarrow$).
}
\label{tab:rollout_length}

\small
\setlength{\tabcolsep}{5pt}

\begin{tabular}{lccc}
\toprule
Rollout Length
& 20
& 40
& 60 \\
\midrule

FNO
& 0.0038
& 0.0130
& 0.0517 \\

FNO+ALR
& \textbf{0.0027}
& \textbf{0.0035}
& \textbf{0.0085} \\

\bottomrule
\end{tabular}

\vspace{-6pt}
\end{wraptable}


We further evaluate FNO and FNO+ALR on Shallow Water with rollout lengths of 20, 40, and 60, with each setting trained and evaluated at the corresponding horizon.
As shown in Table~\ref{tab:rollout_length}, the FNO error increases from $0.0038$ at 20 steps to $0.0130$ and $0.0517$ at 40 and 60 steps, whereas FNO+ALR increases from $0.0027$ to only $0.0035$ and $0.0085$.
Relative to 20 steps, this corresponds to a $13.6\times$ increase for FNO at 60 steps, compared with $3.1\times$ for FNO+ALR.
Figure~\ref{fig:shallow_water_long_rollout} further visualizes the rollout-length-60 setting, where FNO+ALR maintains predictions closer to the ground truth across increasing rollout steps.
Together, the quantitative and qualitative results indicate that cross-origin consistency improves the stability of autoregressive rollout as the rollout length increases.

\section{Conclusion}

We propose Align Latent Rollout (ALR), a regularization approach for improving cross-origin consistency in autoregressive neural PDE solvers.
For a well-posed deterministic PDE flow, states from different temporal origins along the same physical trajectory should evolve to the same shared target state; ALR promotes this property in learned solvers by regularizing their internal representations.
A second-order analysis of latent discrepancy propagation motivates two objectives: zeroth-order latent state consistency and first-order latent dynamics consistency, which align shared-target representations and their local linearized evolution, respectively.
ALR can be integrated into different neural PDE solvers without modifying their backbone architectures.
Experiments across multiple PDE benchmarks and neural-operator backbones consistently show improved prediction accuracy, reduced cross-origin discrepancies, and greater stability under longer autoregressive rollouts.

\newpage

\newpage

\bibliography{iclr2027_conference}
\bibliographystyle{iclr2027_conference}

\newpage

\appendix

\section{Notation}
\label{app:notation}

We distinguish continuous physical time $\tau$ from discrete temporal
indices $t,t_i,t_j$.
Here, $t_i$ denotes a rollout origin, $t_j$ a target time, and
$t_1<t_2<t$ denote two temporal origins and a shared target time.
For rollout-dependent quantities, the subscript
$t_i\rightarrow t_j$ explicitly records the temporal origin and target.

\begin{longtable}{@{}p{0.20\linewidth} p{0.74\linewidth}@{}}
\toprule
\textbf{Symbol} & \textbf{Definition} \\
\midrule
\endfirsthead

\toprule
\textbf{Symbol} & \textbf{Definition} \\
\midrule
\endhead

$\boldsymbol{x}$ &
Spatial coordinate, $\boldsymbol{x}\in\Omega\subset\mathbb{R}^{d}$. \\

$\tau$ &
Continuous physical time, $\tau\in[0,T]$. \\

$t,t_i,t_j$ &
Discrete temporal indices; $t_i$ denotes a rollout origin and
$t_j$ a target time. \\

$t_1,t_2,t$ &
Two temporal origins and a shared target time satisfying
$t_1<t_2<t$. \\

$N$ &
Final discrete temporal index of a trajectory,
$\{\boldsymbol{u}_t\}_{t=0}^{N}$. \\

$\Delta\tau$ &
Temporal discretization interval, with
$\boldsymbol{u}_t=\boldsymbol{u}(\cdot,t\Delta\tau)$. \\

$\boldsymbol{u}(\boldsymbol{x},\tau)$ &
Continuous PDE solution field. \\

$\boldsymbol{u}_t$ &
Ground-truth physical state at discrete time $t$. \\

$\boldsymbol{\mu}$ &
Governing configuration, including PDE coefficients,
external forcing, and boundary data. \\

$\Phi_{t_i\rightarrow t_j}$ &
Exact PDE evolution operator from time $t_i$ to $t_j$:
$\boldsymbol{u}_{t_j}
=
\Phi_{t_i\rightarrow t_j}(\boldsymbol{u}_{t_i})$. \\

$T_{\mathrm{in}}$ &
Input temporal-window length. \\

$T_{\mathrm{out}}$ &
Output temporal-window length. \\

$\boldsymbol{U}_t$ &
Input temporal window ending at $t$:
$\boldsymbol{U}_t=
(\boldsymbol{u}_{t-T_{\mathrm{in}}+1},\ldots,\boldsymbol{u}_t)$. \\

$\boldsymbol{V}_{t+1}$ &
Output temporal window starting at $t+1$:
$\boldsymbol{V}_{t+1}=
(\boldsymbol{u}_{t+1},\ldots,\boldsymbol{u}_{t+T_{\mathrm{out}}})$. \\

$f_\theta$ &
Learned one-step neural operator for PDE evolution. \\

$h$ &
Autoregressive rollout horizon, $h\in\mathbb{Z}_{>0}$. \\

$f_\theta^{(h)}$ &
$h$-fold composition of $f_\theta$:
$f_\theta^{(h)}
=
\underbrace{f_\theta\circ\cdots\circ f_\theta}_{h\ \mathrm{times}}$. \\

$\hat{\boldsymbol{u}}_{t_i\rightarrow t_j}$ &
Autoregressive prediction at target time $t_j$ initialized from
the ground-truth state $\boldsymbol{u}_{t_i}$:
$\hat{\boldsymbol{u}}_{t_i\rightarrow t_j}
=
f_\theta^{(t_j-t_i)}(\boldsymbol{u}_{t_i})$. \\

$K$ &
Consistency-window length used to construct cross-origin pairs
through an auxiliary local autoregressive rollout. \\

$E_\theta$ &
Mapping from the physical input to the selected latent state. \\

$D_\theta$ &
Mapping from the selected latent state to the physical prediction. \\

$\boldsymbol{z}_{t_i\rightarrow t_j}$ &
Selected latent representation associated with target time $t_j$
along the rollout originating at $t_i$. \\

$G_\theta$ &
Induced latent evolution map,
$G_\theta=E_\theta\circ D_\theta$. \\

$\Delta\boldsymbol{z}_t$ &
Cross-origin latent discrepancy at the shared target time:
$\Delta\boldsymbol{z}_t
=
\boldsymbol{z}_{t_1\rightarrow t}
-
\boldsymbol{z}_{t_2\rightarrow t}$. \\

$J_{G_\theta}(\boldsymbol{z})$ &
Jacobian of $G_\theta$ evaluated at latent state $\boldsymbol{z}$. \\

$H_{G_\theta}(\boldsymbol{z})$ &
Second derivative of $G_\theta$ evaluated at latent state
$\boldsymbol{z}$. \\

$\boldsymbol{v}$ &
Stochastic probing direction for Jacobian--vector products,
$\boldsymbol{v}\sim\mathcal{N}(\boldsymbol{0},\mathbf{I})$. \\

$\ell(\cdot,\cdot)$ &
Base physical-state prediction loss. \\

$\mathcal{L}_{\mathrm{pred}}$ &
Original physical-state prediction objective. \\

$\mathcal{L}_{\mathrm{state}}$ &
Zeroth-order latent state consistency loss. \\

$\mathcal{L}_{\mathrm{dynamic}}^{\mathrm{full}}$ &
Full-Jacobian latent dynamics consistency loss. \\

$\mathcal{L}_{\mathrm{dynamic}}$ &
JVP-based first-order latent dynamics consistency loss. \\

$\mathcal{L}_{\mathrm{total}}$ &
Overall training objective. \\

$\lambda_{\mathrm{state}},
\lambda_{\mathrm{dynamic}}$ &
Weights of the state and dynamics consistency objectives. \\

\bottomrule
\end{longtable}

\section{Dataset}

\subsection{Navier--Stokes Equation}
\label{app:ns}

The two-dimensional Navier--Stokes equation considered in the FNO benchmark~\citep{li2020fno}
describes a viscous incompressible fluid in vorticity form on the unit torus.
The governing equations are:
\begin{equation}
\begin{aligned}
\partial_t w(\boldsymbol{x},t)
+ \boldsymbol{u}(\boldsymbol{x},t)\cdot\nabla w(\boldsymbol{x},t)
&=
\nu \Delta w(\boldsymbol{x},t)+f(\boldsymbol{x}),
&& \boldsymbol{x}\in(0,1)^2,\ t\in(0,T],\\
\nabla\cdot \boldsymbol{u}(\boldsymbol{x},t)
&=0,
&& \boldsymbol{x}\in(0,1)^2,\ t\in[0,T],\\
w(\boldsymbol{x},0)
&=
w_0(\boldsymbol{x}),
&& \boldsymbol{x}\in(0,1)^2.
\end{aligned}
\label{eq:ns}
\end{equation}

where $\boldsymbol{x}=(x_1,x_2)$ denotes the spatial coordinate and
$t$ denotes time.
The velocity field is represented by
$\boldsymbol{u}(\boldsymbol{x},t)$, and
$w(\boldsymbol{x},t)=\nabla\times\boldsymbol{u}(\boldsymbol{x},t)$
denotes the corresponding vorticity field.
The initial vorticity is denoted by $w_0(\boldsymbol{x})$.
Here, $\nu>0$ is the kinematic viscosity coefficient,
$f(\boldsymbol{x})$ is the external forcing term,
$\nabla$ denotes the spatial gradient operator,
and $\Delta$ denotes the Laplacian operator.
The incompressibility condition is imposed through
$\nabla\cdot\boldsymbol{u}=0$.
Periodic boundary conditions are applied on the spatial domain $(0,1)^2$.


\subsection{Shallow-Water Equation}
\label{app:swe}

The two-dimensional shallow-water equations (SWE) considered in PDEBench~\citep{takamoto2022pdebench}
describe the evolution of free-surface flows in conservative form.
The governing equations are:
\begin{equation}
\begin{aligned}
\partial_t h
+\partial_x(hu)
+\partial_y(hv)
&=0,\\
\partial_t(hu)
+\partial_x\left(hu^2+\frac{1}{2}gh^2\right)
+\partial_y(huv)
&=-gh\,\partial_x b,\\
\partial_t(hv)
+\partial_x(huv)
+\partial_y\left(hv^2+\frac{1}{2}gh^2\right)
&=-gh\,\partial_y b .
\end{aligned}
\label{eq:swe}
\end{equation}

where $\boldsymbol{x}=(x,y)$ denotes the spatial coordinate and $t$ denotes time.
The water depth is represented by $h(\boldsymbol{x},t)$, while
$u(\boldsymbol{x},t)$ and $v(\boldsymbol{x},t)$ denote the two velocity components of the velocity field:
\[
\boldsymbol{u}(\boldsymbol{x},t)
=
\begin{pmatrix}
u(\boldsymbol{x},t)\\
v(\boldsymbol{x},t)
\end{pmatrix}.
\]
The quantities $hu$ and $hv$ represent the corresponding momentum components,
$b(\boldsymbol{x})$ denotes the bathymetry function,
and $g$ is the gravitational acceleration.

The PDEBench benchmark considers a two-dimensional radial dam-break problem
on the square domain $\Omega=[-2.5,2.5]^2$.
The initial water depth is defined as:
\begin{equation}
h(\boldsymbol{x},0)=
\begin{cases}
2.0, & \|\boldsymbol{x}\|_2<r,\\
1.0, & \|\boldsymbol{x}\|_2\ge r,
\end{cases}
\label{eq:swe_initial}
\end{equation}
where $r$ denotes the radius of the initial circular region and is sampled from
the uniform distribution $\mathcal{U}(0.3,0.7)$ for each trajectory.


\subsection{Kolmogorov Flow}
\label{app:kolmogorov}

The two-dimensional Kolmogorov flow dataset is generated from the incompressible
Navier--Stokes equation in vorticity form.
The governing equations are:
\begin{equation}
\begin{aligned}
\partial_t \omega(\boldsymbol{x},t)
+\boldsymbol{u}(\boldsymbol{x},t)\cdot\nabla\omega(\boldsymbol{x},t)
&=
\frac{1}{Re}\Delta\omega(\boldsymbol{x},t)
+f(\boldsymbol{x},t),\\
\nabla\cdot\boldsymbol{u}(\boldsymbol{x},t)
&=0,\\
\omega(\boldsymbol{x},0)
&=\omega_0(\boldsymbol{x}),
\end{aligned}
\label{eq:kolmogorov}
\end{equation}
where $\boldsymbol{x}=(x_1,x_2)$ denotes the spatial coordinate,
$t$ denotes time,
$\omega(\boldsymbol{x},t)$ represents the vorticity field,
and $\boldsymbol{u}(\boldsymbol{x},t)$ denotes the corresponding velocity field.
$Re$ represents the Reynolds number,
$f(\boldsymbol{x},t)$ denotes the forcing term,
$\nabla$ denotes the spatial gradient operator,
and $\Delta$ denotes the Laplacian operator.

The forcing term is defined as:
\begin{equation}
f(\boldsymbol{x},t)
=
-n\cos(nx_2)-0.1\omega(\boldsymbol{x},t),
\label{eq:kolmogorov_forcing}
\end{equation}
where $n$ controls the forcing frequency.
Following the settings in previous PDE surrogate modeling studies
~\citep{li2023scalableTransformer},
the forcing frequency is set to $n=8$.
The additional linear drag term
$-0.1\omega(\boldsymbol{x},t)$ is introduced to model dissipative effects
following Kochkov et al.~\citep{kochkov2021machine}.

The system is defined on the periodic domain
$\Omega=(0,2\pi)^2$ with periodic boundary conditions in all spatial directions.
The initial vorticity field
$\omega_0(\boldsymbol{x})$
is sampled from a prescribed Gaussian random field following the setting of
Li et al.~\citep{li2020fno}.





\section{Baseline Neural Solvers}
\label{app:baseline_solvers}

\textbf{Fourier Neural Operator (FNO).}
We use a four-layer Fourier Neural Operator (FNO)~\citep{li2020fno} on structured
two-dimensional grids. The input field and spatial coordinates are first mapped to
a 64-channel feature representation through a pointwise lifting module. The backbone
then applies four Fourier operator layers, each using 12 retained Fourier modes along
each spatial dimension and combining a spectral convolution with a pointwise
transformation. GELU activation is applied after the first three Fourier layers.
The final feature representation is mapped to the physical output through a two-layer
pointwise projection with an intermediate GELU activation.
For ALR, we select the activated output of the second Fourier layer as the latent
representation. The lifting module and the first two Fourier layers therefore form
the representation-extraction part, while the remaining two Fourier layers and the
output projection form the subsequent prediction mapping.

\textbf{Factorized Fourier Neural Operator (F-FNO).}
We use the Factorized Fourier Neural Operator
(F-FNO)~\citep{tran2023factorized} on structured two-dimensional grids.
The input field and spatial coordinates are first mapped to a 64-channel
feature representation through a pointwise lifting module.
The lifted features are then processed by a sequence of residual factorized
spectral blocks, where the Fourier transform is applied separately along
the spatial dimensions.
Each block updates the feature representation through a residual connection.
The resulting features are mapped to the physical output through a two-layer
pointwise projection with an intermediate GELU activation.
For ALR, we select the feature after the second residual
factorized spectral block as the latent representation.
The subsequent spectral blocks and output projection retain the original
F-FNO structure.

\textbf{Latent Spectral Model (LSM).}
We use the Latent Spectral Model (LSM)~\citep{wu2023lsm} on structured
two-dimensional grids, with base channel width 64 and eight attention heads.
In our implementation, the input is first lifted to a 64-channel feature representation.
The hierarchical encoder consists of an initial convolutional block followed by four downsampling stages, producing feature maps at five spatial resolutions.
A neural spectral block is associated with each resolution level.
The decoding path contains four upsampling stages with skip connections from the corresponding encoder levels, followed by an output convolution and a two-layer pointwise projection with a GELU activation.
For ALR, we select the output of the fourth downsampling stage, immediately before
the coarsest-scale neural spectral block, as the latent representation. The remaining
neural spectral blocks, upsampling path, skip connections, and output projection are
kept unchanged.

\section{Experiment Details}

\subsection{Dataset Configuration}

\textbf{Navier--Stokes ($\nu=10^{-5}$).}
The Navier--Stokes benchmark is taken from the public dataset released with
the Fourier Neural Operator (FNO)~\citep{li2020fno}.
It contains trajectories of the two-dimensional incompressible Navier--Stokes
equations in vorticity form on a periodic spatial domain.
The original numerical solutions were generated on a $256\times256$ uniform
grid and subsequently subsampled to $64\times64$ for the standard FNO benchmark.

For $\nu=10^{-5}$, the public Navier--Stokes dataset contains 1,200
trajectories, each comprising 20 temporal states on a $64\times64$ spatial grid with a single vorticity channel.
The first 800 trajectories are used for training, the subsequent 200 for
validation, and the final 200 for testing.
The temporal history window is fixed to one physical state.
Starting from the initial physical state, the solver advances autoregressively for 19 successive temporal steps to cover the remaining trajectory.
No additional spatial downsampling or data normalization is applied.

\textbf{Navier--Stokes ($\nu=10^{-4}$).}
The dataset is taken from the public Navier--Stokes benchmark released with
FNO~\citep{li2020fno}, consisting of temporal trajectories of the
two-dimensional incompressible Navier--Stokes equations in vorticity form on
a periodic spatial domain.
For the viscosity setting $\nu=10^{-4}$, the public
Navier--Stokes dataset contains 10,000 trajectories,
each comprising 30 temporal physical states of a single-channel vorticity
field on a $64\times64$ spatial grid.

The experiments use the first 800 trajectories for training, the subsequent
100 for validation, and the following 100 for testing.
The temporal history window is fixed to one physical state.
Starting from the initial physical state, the solver advances autoregressively
for 29 temporal steps over the remaining trajectory.
The original spatial resolution is retained without additional downsampling,
and no data normalization is applied.

\textbf{Shallow Water.}
The Shallow Water dataset is taken from the two-dimensional radial dam-break benchmark in PDEBench~\citep{takamoto2022pdebench}.
The simulations are defined on a square spatial domain and describe the temporal evolution of a single-channel water-depth field $h$ from randomized radial dam-break initial conditions.
The public dataset contains 1,000 trajectories at a spatial resolution of$128\times128$.

The first 800 trajectories are used for training, the subsequent 100 for
validation, and the final 100 for testing.
A spatial downsampling factor of two is applied along both spatial dimensions, resulting in a $64\times64$ grid, while the original temporal resolution is retained.
The water-depth field is standardized using a scalar mean and standard deviation computed exclusively from the training split.
The temporal history window is fixed to one physical state.
The standard experimental setting uses a 20-step autoregressive rollout horizon. 
For the rollout-length analysis in Section~\ref{sec:rollout_length}, additional 40-step and 60-step settings are considered.

\textbf{Kolmogorov Flow.}
The Kolmogorov Flow benchmark follows the two-dimensional turbulent-flow
configuration considered by Kochkov et al.~\citep{kochkov2021machine}.
The system is governed by the incompressible Navier--Stokes equations at
$\mathrm{Re}=1000$ with Kolmogorov forcing, producing statistically stationary turbulent dynamics.

The data used in the experiments contain 120 trajectories, each comprising 320 temporal physical states of a single-channel scalar flow field on a $256\times256$ spatial grid.
The first 100 trajectories are used for training, followed by 10 trajectories for validation and 10 for testing.
For the reported experiments, the spatial fields are subsampled to a
$128\times128$ grid, while the original temporal resolution is retained.
The scalar field is standardized using a mean and standard deviation computed exclusively from the training split.
The temporal history window is fixed to one physical state, and the solver is evaluated over a 20-step autoregressive rollout horizon.

\subsection{Training Configuration}
\label{app:training_configuration}

All models are trained for 500 epochs using AdamW with a learning rate of $5\times10^{-4}$, weight decay $10^{-5}$, and a OneCycle learning-rate schedule.
The relative $\ell_2$ error is used as the prediction objective.
A batch size of 20 is used for both Navier--Stokes and Shallow Water, while Kolmogorov Flow uses a batch size of 5 due to its higher spatial resolution.
The standard experiments adopt ground-truth-conditioned one-step prediction as the primary training objective.
For ALR, one temporal origin is sampled independently for each trajectory, and an auxiliary local autoregressive rollout is used only to construct cross-origin consistency pairs; no additional prediction loss is imposed on this auxiliary rollout.
Unless otherwise specified, the consistency window is fixed to $K=6$, and valid temporal origins are sampled uniformly.
The state and dynamics consistency terms are both weighted by $0.1$.
For the Navier--Stokes experiments, state consistency is activated after 50 epochs and dynamics consistency after 100 epochs.
For Shallow Water and Kolmogorov Flow, the state consistency and latent dynamics consistency warm-up periods are 100 and 200 epochs, respectively.

\subsection{Evaluation Metric}
\label{app:metric}

We use the relative $\ell_2$ error as the evaluation metric:
\begin{equation}
\mathcal{E}_{\mathrm{rel}}
=
\frac{
\left\|
\boldsymbol{u}-\hat{\boldsymbol{u}}
\right\|_2
}{
\left\|
\boldsymbol{u}
\right\|_2
},
\end{equation}
where $\boldsymbol{u}$ and $\hat{\boldsymbol{u}}$ denote the ground-truth and predicted solutions, respectively. Lower values indicate more accurate predictions.

\section{Computational Efficiency}
\label{subsec:efficiency}

To assess the computational overhead of ALR, additional efficiency experiments
are conducted on Navier--Stokes ($\nu=10^{-5}$) with different
consistency-window lengths $K$.
Table~\ref{tab:efficiency} reports the prediction error together with the
normalized training and inference time.
The first row corresponds to the original FNO baseline, whose runtime is
normalized to $1.00\times$; all other runtime values are reported relative to
this baseline.

ALR introduces additional computation only during training, where auxiliary
local rollouts are constructed to optimize the cross-origin consistency
objectives.
Consequently, the training cost increases with $K$, from $2.04\times$ at
$K=4$ to $4.04\times$ at $K=10$.
In contrast, the auxiliary rollout and consistency objectives are removed
during inference, leaving the original FNO prediction procedure unchanged.
Therefore, ALR introduces no additional inference-time overhead, with the
normalized inference time remaining $1.00\times$ for all settings.
In terms of accuracy, increasing $K$ from 4 to 6 reduces the relative
$\ell_2$ error from $0.0893$ to $0.0858$, while further increasing $K$
produces only marginal changes despite noticeably higher training cost.
Accordingly, $K=6$ is used as the default setting, providing a practical
trade-off between prediction accuracy and training efficiency.

\begin{table}[!htbp]
\centering
\caption{Computational efficiency under different consistency-window lengths
$K$ on Navier--Stokes ($\nu=10^{-5}$).
Training and inference times are normalized by the FNO baseline.
Lower relative $\ell_2$ error and runtime are better.}
\label{tab:efficiency}

\vspace{5pt}

\begin{tabular}{cccc}
\toprule
$K$
& Relative $\ell_2$ $\downarrow$
& Training Time ($\times$) $\downarrow$
& Inference Time ($\times$) \\
\midrule
--  & 0.1004          & 1.00 & 1.00 \\
4   & 0.0893          & 2.04 & 1.00 \\
6   & 0.0858          & 2.66 & 1.00 \\
8   & \textbf{0.0851} & 3.42 & 1.00 \\
10  & 0.0856          & 4.04 & 1.00 \\
\bottomrule
\end{tabular}
\end{table}

\section{Qualitative Visualization of Autoregressive Rollout}
\label{app:ns1e4_rollout_vis}

\begin{figure*}[t]
    \centering
    \includegraphics[width=\textwidth]{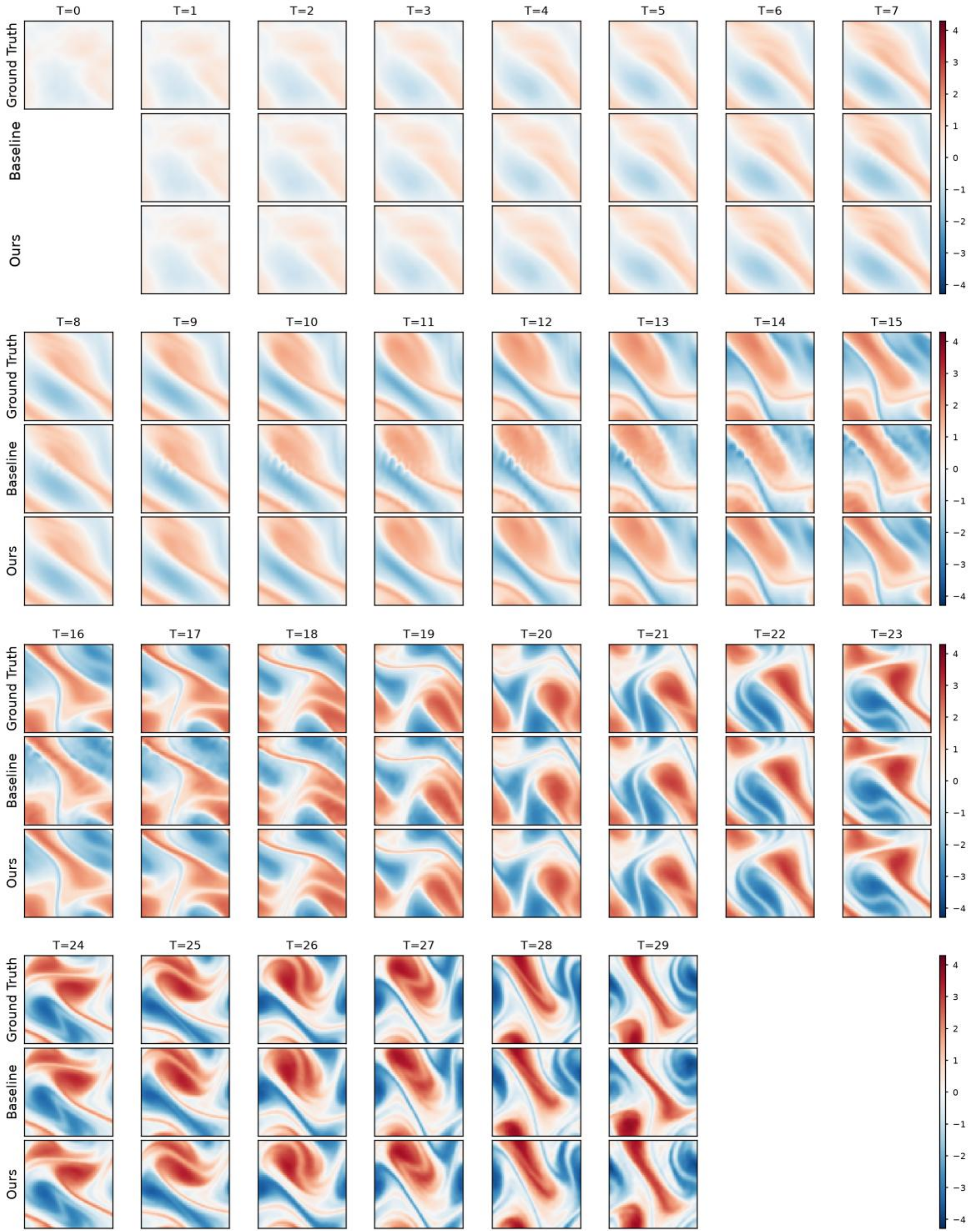}
    \vspace{-6pt}
    \caption{
    Qualitative comparison of the complete 29-step autoregressive rollout on
    Navier--Stokes ($\nu=10^{-4}$).
    The initial physical state is shown at $T=0$.
    For subsequent temporal steps, ground truth, FNO baseline, and FNO+ALR
    predictions are shown from top to bottom.
    All models start from the same initial state and are rolled out
    autoregressively over the full temporal horizon.
    }
    \label{fig:ns1e4_full_rollout}
\end{figure*}

Figure~\ref{fig:ns1e4_full_rollout} visualizes the complete autoregressive
rollout on Navier--Stokes ($\nu=10^{-4}$).
Starting from the same initial physical state at $T=0$, both the FNO baseline
and FNO+ALR are rolled out autoregressively for 29 successive temporal steps.
The ground-truth trajectory is shown together with the corresponding predictions
from the two models at each target time.

At the early rollout stages ($T=1$--$8$), both methods remain close to the
ground-truth evolution.
Visible differences begin to emerge for the FNO baseline around
$T=9$--$10$, particularly near the central transition between positive and
negative vorticity regions, where small phase shifts and boundary distortions
start to appear.
These discrepancies become more pronounced around $T=13$--$15$, as the
location and curvature of the evolving structures increasingly deviate from
the ground truth.
At later stages ($T\geq16$), the accumulated rollout error leads to clearer
misalignment in the large-scale flow structures and their temporal evolution.
In contrast, FNO+ALR remains more closely aligned with the ground-truth
trajectory across these intermediate and late rollout stages, preserving the
position, shape, and evolution of the dominant vorticity structures more
consistently.
This qualitative comparison illustrates that the benefit of ALR becomes more
apparent as the rollout depth increases, consistent with its objective of
reducing the error accumulation during autoregressive
evolution.

\end{document}